\documentclass[applsci,article,accept,pdftex,moreauthors]{Definitions/mdpi} 
\firstpage{1} 
\pubvolume{1}
\issuenum{1}
\articlenumber{0}
\pubyear{2024}
\copyrightyear{2024}
\externaleditor{~}
\datereceived{20 December 2024} 
\daterevised{10 February 2025} 
\dateaccepted{11 February 2025} 
\datepublished{ } 
\hreflink{https://doi.org/} 

\usepackage{algorithm}
\usepackage{algpseudocode}
\usepackage{graphicx}
\usepackage{multirow}
\Title{Lightweight Deepfake Detection Based on Multi-Feature Fusion} 

\TitleCitation{Lightweight Deepfake Detection Based on Multi-Feature Fusion}

\Author{Siddiqui Muhammad Yasir and Hyun Kim *\orcidA{}}

\AuthorNames{Siddiqui Muhammad Yasir and Hyun Kim}

\AuthorCitation{Yasir, S.M.; Kim, H.}

\address[1]{%
Department of Electrical and Information Engineering, Research Center for Electrical and Information Technology, Seoul National University of Science and Technology, 232 Gongneung-ro, Nowon-gu, \mbox{Seoul 01811, Republic of Korea;} siddiqui@seoultech.ac.kr}

\corres{\hangafter=1 \hangindent=1.05em \hspace{-0.82em} Correspondence: hyunkim@seoultech.ac.kr}

\abstract{Deepfake technology utilizes deep learning (DL)-based face manipulation techniques to seamlessly replace faces in videos, creating highly realistic but artificially generated content. Although this technology has beneficial applications in media and entertainment, misuse of its capabilities may lead to serious risks, including identity theft, cyberbullying, and false information. The integration of DL with visual cognition has resulted in important technological improvements, particularly in addressing privacy risks caused by artificially generated ``deepfake'' images on digital media platforms. In this study, we propose an efficient and lightweight method for detecting deepfake images and videos, making it suitable for devices with limited computational resources. In order to reduce the computational burden usually associated with DL models, our method integrates machine learning classifiers in combination with keyframing approaches and texture analysis. Moreover, the features extracted with a histogram of oriented gradients (HOG), local binary pattern (LBP), and KAZE bands were integrated to evaluate using random forest, extreme gradient boosting, extra trees, and support vector classifier algorithms. Our findings show a feature-level fusion of HOG, LBP, and KAZE features improves accuracy to 92\% and 96\% on FaceForensics++ and Celeb-DF(v2), respectively.}


\begin{document}

\section{Introduction}
The development of deepfake technologies presents significant challenges for visual cognition in deep learning (DL) and raises serious concerns for visual information risks, as these synthetic media convincingly manipulate visual content, spreading misinformation~\mbox{\cite{10552098, info15090525, Bale2024}}. The manipulated content may raise concerns about the potential abuse of technology and its consequences on politics, finance, and personal privacy~\cite{info15090525, 10307811}. Although convolutional neural networks (CNNs) have achieved considerable success in computer vision tasks~\cite{idsl_embedded2,idsl_performance2,idsl_performance4,idsl_performance5} including deepfake detection, alternative approaches that combine machine learning and hand-made features are gaining popularity. In computer vision, technologies such as auto-encoders and generative adversarial networks (GANs) facilitate the generation of manipulated visuals. Deepfake images are often categorized into facial synthesis, attribute manipulation, identity swapping, and expression swapping, with CNN models commonly employed for video detection~\cite{10307811, ABBAS2024124260}.
To identify highly precise synthetic visual data as deepfake, an effective deepfake detection method is required~\cite{info15090525, NASKAR2024e25933}. DL reduces human effort in feature engineering but increases complexity and interpretability due to high nonlinearity and input interactions. Traditional machine learning (ML) methods often sacrifice accuracy for interpretability due to their complexity and large data volumes. DL methods are challenging to train and require computing resources, while ML methods are easier to evaluate and understand~\cite{9790940, app12178455}. To address these constraints, we encouraged to experiment and evaluate traditional machine learning techniques to detect deepfakes.

Machine learning (ML) classifiers that use features such as local binary pattern (LBP)~\cite{MOORE2011541} and KAZE descriptor~\cite{10.1007/978-3-642-33783-3_16} offer promising alternatives to CNN-based methods for deepfake detection. LBP encodes texture information through local spatial patterns~\cite{MOORE2011541}, while KAZE provides robust, noise-invariant descriptors~\cite{10.1007/978-3-642-33783-3_16} and key points, allowing the detection of minor deepfake artifacts that CNNs may overlook~\cite{app12178455, MOORE2011541, 10.1007/978-3-642-33783-3_16}.~Traditional models can be improved in accuracy and resilience by integrating machine learning feature extraction methods~\cite{Huda2024}. Smaller datasets can benefit greatly from the use of LBP and KAZE features, which increase detection process transparency~\cite{AKhalid2024, Ghosh2022}. To address the growing difficulties presented by deepfake technology, our hybrid approach—which integrates feature-based classifiers—offers a potential path ahead in differentiating real from fake visual data~\cite{10354308}.

Deepfake detection solutions are limited for social media analysis due to heavy compression~\cite{Chen2022}. A compact model, based on ML classifiers, is needed for memory-constrained devices like smartphones~\cite{idsl_hardware1,idsl_hardware2}. The proposed model, focusing on auto-encoder-based generated videos and keyframe identification, achieves high accuracy with minimal computational demands, making it suitable for memory-constrained devices and enhancing deepfake detection capabilities~\cite{Du2020, 10678555}.

This study aims to reduce computational costs in deepfake detection without substantial accuracy loss. Our proposed model targets compressed social media videos, focusing on the existing multi-feature fusion approach using multiple feature types (HOG, LBP, and KAZE) for a more comprehensive representation of visual data. By analyzing variations in visual artifacts, we achieve significant data reduction while preserving accuracy. The method integrates well-known ML feature extraction methods with diverse texture features, allowing effective training on limited datasets. The top three classifiers are presented in Table~
 \ref{tab:fusion-LBP-KAZE-features}. We evaluated the proposed fusion model using the Face Forensic++~\cite{roessler2019faceforensicspp} and Celeb-DF~\cite{Li_2020_CVPR} dataset, which replicates scenarios commonly found on social media platforms. The use of FaceForensics++ and Celeb-DF datasets aligns with the common practice in the field, allowing for a direct comparison with existing methods. The main contributions of this work are summarized as follows:

\begin{itemize}
	\item The proposed fusion model introduces a novel approach to deepfake detection on platforms with limited memory and processing capabilities, effectively managing compressed video data;
	\item Using existing classification techniques for artifact analysis, the method achieves substantial data reduction while preserving detection accuracy;
	\item The methodology combines forty established ML classifiers (using HOG, LBP, and KAZE features) with diverse texture-based features, demonstrating reliable performance even with limited datasets;
	\item The evaluation primarily uses the Face Forensic++ dataset, which reflects real-world scenarios and emphasizes minimizing computational overhead.
\end{itemize}

The remainder of this paper is organized as follows. Section~\ref{sec:impact} examines the impact of deepfake technology across different platforms. Section~\ref{sec:proposed-framework} reviews lightweight feature detection methods and details the architecture of the proposed fusion model. Section~\ref{sec:exp-result} describes the dataset, experimental setup, and evaluation metrics used to assess the proposed fusion model, as well as its limitations. Finally, Section~\ref{sec:conclusion} summarizes the study findings and outlines potential future directions.

\section{Related Works}
\label{sec:impact}

Deepfakes have emerged as a critical challenge, prompting extensive research into detection techniques. The DL-based methods have shown the most advancement, leading to efficient detection systems. While various approaches have been proposed, they primarily rely on similar underlying principles~\cite{idsl_survey1,https://doi.org/10.1002/widm.1520}. 
Most of the detection methods use CNN-based models to classify images as fake or real, but state-of-the-art deepfake detectors (e.g., N. Bonettini~\cite{9412711}) still rely on complex neural networks, struggle with generalization to unseen deepfake techniques, and lack robustness under real-world distortions~\cite{9412711, saberi2024robustness}.

Several deepfake detection approaches depend on various modalities and feature fusion to improve accuracy. Prior research has shown that integrating spatial and frequency domain features, as well as combining spatial, temporal, and spatiotemporal features, significantly improves detection accuracy compared to single-modality approaches~\cite{dong2023contrastive, raza2023holisticdfd, zhu2024high}. For instance, Almestekawy et~al.~\cite{ALMESTEKAWY2024100535} fused Facial Region Feature Descriptor (FFR-FD) with random forest classifier and texture features (standard deviation, gradient domain, and GLCM) fed into an SVM classifier. Raza et~al.~\cite{raza2023holisticdfd} proposed a three-stream network utilizing temporal, spatial, and spatiotemporal features for deepfake detection. Moreover, security techniques for deepfake detection on untrusted servers were introduced by Chen B. et~al.~\cite{chen2023privacy}; their method enables distant servers to detect deepfake videos without understanding the~\mbox{content}.

Proper methods are essential for extracting valuable information from large unprocessed visual data, with feature-based techniques like LBP and KAZE offering computational efficiency as an alternative to resource-intensive CNNs~\cite{app12178455}. Recent studies have suggested that combining extracted features with advanced ML classifiers can develop hybrid models for deepfake detection while maintaining robustness across diverse datasets~\cite{Huda2024, app12178455, 8346440}. 

Alternatively, texture can be encoded by comparing each pixel with its neighbors, creating a binary pattern that serves as a robust feature descriptor across various lighting conditions. Feature extraction techniques are divided into global and local descriptor approaches~\cite{9760679}. Global methods analyze the entire image to generate a feature vector and are considered fast processing but have some limitations, such as Principle Component Analysis (PCA)~\cite{CAVALCANTI20134971}, Linear Discriminant Analysis (LDA)~\cite{lu2012incremental}, and Global Gabor generic features~\cite{fathi2016new}. Local descriptors, like LBP~\cite{topi2000robust} and Histogram of Oriented Gradients (HOG)~\cite{dalal2005histograms}, provide a more effective representation of images. LBP is widely used in face recognition~\cite{deniz2011face}, while HOG is used for human detection by dividing the image into fixed-size blocks and computing HOG features for each block. Likewise, the selection of custom features (Local Binary Patterns (LBP) based on texture and a customized High-Resolution Network (HRNet)) proposed by Khalil et~al.~\cite{khalil2021icaps} and fed to the SVM classifier. 
This efficiency makes LBP a popular choice in tasks where texture details are important, such as facial recognition and expression analysis, while also reducing processing time and computational costs~\cite{app14156567}. Deepfake artifacts regularly change gradient orientations and edge patterns, which are essential for lightweight detection on resource-constrained devices. Compared to CNN-based approaches, it is less successful in detecting higher-level semantic discrepancies~\cite{ASLAN2020107704}. KAZE, on the other hand, can detect unique key points that are invariant to noise and transformations, which is essential for applications requiring high-fidelity feature matching under variable conditions. By detecting and characterizing two-dimensional features in nonlinear scale space, the KAZE features~\cite{10.1007/978-3-642-33783-3_16} resist Gaussian blurring. KAZE’s reliance on nonlinear diffusion allows it to capture image structures that are often missed by traditional linear approaches, enhancing performance in complex environments~\cite{8464688}.

More recent deepfake methods, particularly diffusion models, have introduced high-quality synthetic images that closely resemble natural visuals, evading common detection markers such as GAN-related grid artifacts~\cite{Zotova2024}. Chen Y et~al.~and Yuan et~al.~\cite{chen2023text, yuan2023inserting} developed a model that uses a reference image and text prompt to generate deepfake images as human identity. These developments initiate a shift in detection strategies, where integrating extracted features with classifiers holds significant potential for improving accuracy and reducing computational load~\cite{ALMESTEKAWY2024100535, https://doi.org/10.1002/widm.1520, NASKAR2024e25933, Abhisheka_2023, Mohtavipour_2021}. 

As a result, detecting deepfake images/videos contributes to the struggle against spreading false information and encourages preserving the validity of visual content and privacy. Our methodology differs from previous approaches in numerous important ways, including the use of multi-feature-level fusion (HOG, LBP, and KAZE features) prior to classification. Focus on characteristics that are computationally efficient. For validation, supervised ML classifiers (such as support vector machines (SVMs), random forest (RF), and gradient boosting classifiers) were used, and their performance in deepfake detection has been evaluated.

\section{Proposed Fusion Model}
\label{sec:proposed-framework}
In machine learning, a feature refers to a specific, measurable attribute of an image that helps in distinguishing patterns. This study focuses on the integration of two types of features (local descriptors) obtained from the LBP and HOG with KAZE features before the classification. To reduce computing costs and meaningful results, the detection of important frames and the elimination of insufficient frames are necessary. In the first step, the keyframes are extracted from videos within an interval of 0.5 s. 
 The first and last 10 frames are overlooked (only if necessary) because they usually have information about the introduction or credits that are not directly related to deepfake detection preprocessing. Additionally, to identify the important keyframes, a similarity check between frames is used as the criterion. Various threshold values are used to determine different approaches. This part of the algorithm results in a pool of distinctive images of frames, which are available for feature extraction. The images were resized to a 28 × 28 single-channel format, ensuring a standardized input for processing. The research explores the use of frames extracted from video footage or standalone images, treating keyframes as textured representations. In the second step of feature extraction, the LBP, HOG, and KASE techniques are applied separately, which typically generates histograms. Furthermore, the fusion of LBP and KAZE and that of HOG and KAZE are used to input futures for the Extra Trees, Random Forest, Support Vector, and XGB classifiers. The proposed fusion model aims to improve detection accuracy while maintaining efficiency.

\begin{figure}[H]
	\includegraphics[scale=0.67]{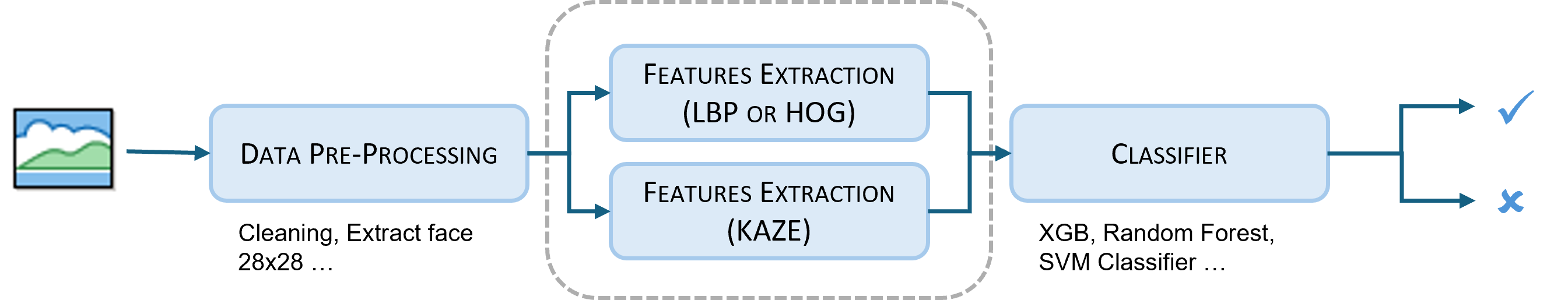}
	\caption{General abstract of the proposed feature-level fusion method.}
	\label{fig:frame-work}
\end{figure}

Figure~\ref{fig:frame-work} illustrates a general abstract of the proposed fusion model with important key concepts involved as data preprocessing (extracting keyframes), extracting features from keyframes using LBP and HOG feature extraction methods, feature-level fusion with KAZE features, classifier selection of the combined features, and subsequent classification on the basis of the chosen classifiers. The keyframes are converted to a logarithmic scale and divided into multiple bands to capture localized information about the texture patterns. We analyze each band of these frame divisions by calculating the normalized histogram of HOG or LBP features to pinpoint characteristics using our classifiers. After these histograms are combined to form an LBP feature vector, they are combined with KAZE features to create supporting feature sets. To handle the dimensionality caused by merging features at the level of characteristics, various classifiers are utilized to identify features, with importance scores that are then fed into the ultimate classifier.

The importance and understanding of feature extraction, along with its configurations, are explained in the following section.

\subsection{LBP Features}
The LBP is an advanced technique for extracting features from images for texture study due to its efficient handling of value variations and straightforward computational process~\cite{Devi_2020}. LBP sets thresholds for neighboring pixels, enabling accurate spatial pattern extraction from images and transforming textual information into binary data for classification and detection~\cite{Werghi_2015}.
LBP analyzes every pixel in an image by evaluating the relationship between each pixel and its surrounding pixels within a specified radius \(R\) distance away from it. If the neighboring pixel value exceeds that of the pixel level, a binary bit is assigned as \(1\); otherwise, it is marked as \(0\)~\cite{Kumar_2022}.

Given a grayscale image \(I \) of size \(M \times N \), the LBP feature for each pixel \((x,y)\) is computed via the following formula:
\begin{equation}
	\text{LBP}(x_p,y_p) =
	\begin{cases}
		1 & \text{if } I(x_p,y_p)  \ge I(x,y) \\
		0 & otherwise
	\end{cases}
\end{equation}

For a pixel \((x,y) \), we compare its intensity \(I(x,y) \) with the intensities of its \(P\) neighboring pixels on a circle of radius \(R\). Let the intensities of these neighbors be \(\{I(x_p,y_p)\}P_{P=1}\). A binary value is assigned to each neighboring pixel.
These binary values are concatenated to form a binary number, which is then converted to a decimal value. Compute the histogram of these LBP values over the entire image.
\[
H_{\text{LBP}}(k) = \sum_{x=1}^{M} \sum_{y=1}^{N} \delta(\text{LBP}(x,y), k), \quad k \in \{0, 1, \ldots, 2^P - 1\}
\]
where \((a,b)\) is the Kronecker delta function, which is \(1\) if \(a=b\) and \(0\) otherwise. Finally, the histogram is normalized with a small constant to prevent division by zero.

The calculation used in this study is explained as follows: for a given pixel \((x_p,y_p) \), the intensity \(I(p_i) \) in the center of the \((3 \times 3) \) block is computed by comparing \(x_p \) to its \(8 \) neighboring pixels. 
The texture classification process relies on illumination, translation, and rotational variance, while keyframes lack control over these attributes, focusing instead on uniform pattern representation. A uniform pattern, \(LBP(x_p,y_p)\), is more suitable. Preliminary experiments showed \(P = 12 \) and \(R = 2 \) provide the best performance for the feature descriptor. With regard to categorizing textures on the basis of their patterns in images or videos, the way the pattern is perceived is influenced by factors such as lighting changes, shifting positions, and orientations of the texture details. However, when we focus on moments in a sequence, the patterns are not affected by rotations or translations. Instead, they are determined by how colors and contrasts spread out over different frequencies and points in time, creating a consistent pattern overall. Therefore, using patterns such as \(LBP(x_p,y_p)\) is more suitable for detecting deepfake keyframes. To evaluate this, we conducted some experiments where we tried different values for the radius \(R \) as well as the neighboring pixel \(P \). After conducting our analysis, we determined that the values of \(P = 12 \) and \(R = 2 \) are suitable for the feature descriptor~\cite{Kumar_2022}.

These comparisons provide a binary vector that represents the connection between the intensity of one pixel and its neighbors. LBPs are widely utilized intensity-based features in various domains, including human detection~\cite{Karunarathne_2019, Yang_2013}, facial recognition~\cite{Karunarathne_2019}, background subtraction, and textured surface recognition~\cite{Huang_2021}. The LBP operator is particularly appealing because of its computational simplicity~\cite{Rahayu_2020}.
However, a notable limitation of the LBP method is the extensive number of histogram bins needed, which reduces its efficiency for localized image patches. Despite this drawback, LBP remains effective for global image representations, making it a valuable tool for image classification tasks.

\subsection{HOG Features}
The HOG is a feature descriptor widely used for texture analysis and object detection~\cite{albiol2008face}. This method is particularly suitable for tasks requiring robust edge and gradient-based analysis, such as detecting structural inconsistencies in deepfake images. HOG divides an image into smaller spatial regions, known as cells, and computes a histogram of gradient directions within each cell. This process can be summarized into three steps: gradient calculation, cell histogram generation, and feature vector construction:

\begin{itemize}
	\item Gradient Calculation: For each pixel in the image, the gradients along the \(x\)- and \(y\)-axes are calculated using Sobel filters:
     \[
     G_x = I(x+1, y) - I(x-1, y), \quad G_y = I(x, y+1) - I(x, y-1)
     \]
     
   The magnitude \(M\) and direction \(\theta\) of the gradient are computed as:
     \[
     M = \sqrt{G_x^2 + G_y^2}, \quad \theta = \arctan\left(\frac{G_y}{G_x}\right)
     \]

\item Cell Histogram Generation: The gradient magnitudes \(M\) are binned into orientation histograms, where the direction \(\theta\) is quantized into a fixed number of bins (e.g., 9 bins for 0°–180° or 18 bins for 0°–360°). To improve invariance to illumination and contrast changes, the histograms are normalized within overlapping spatial blocks. Given a block \(B\), normalization can be performed as:
     \[
     \text{HOG}_{\text{norm}}(B) = \frac{\text{HOG}(B)}{\sqrt{\|\text{HOG}(B)\|^2 + \epsilon}}
     \]
     where \(\epsilon\) is a small constant to prevent division by zero.

\item Feature Vector Construction: The normalized histograms obtained from all the blocks are concatenated to form a single feature vector representing the image. HOG captures fine-grained details about edge orientations and their distribution, making it suitable for identifying subtle spatial distortions caused by deepfake manipulations.
\end{itemize}

In this study, the following HOG parameters were used:

\begin{itemize}
	\item  Cell Size: \(8 \times 8\) pixels;
\item Block Size: \(2 \times 2\) cells;
\item Number of Orientation Bins: 9 (0°–180°);
\item Step Size: \(50\%\) overlap between blocks. 
\end{itemize}

HOG is an effective choice for resource-constrained deepfake detection due to these parameters, which maintain a balance between descriptive strength and processing efficiency. HOG can be used in combination with other robust feature descriptors, such as KAZE, to improve its sensitivity to high-level semantic adjustments to improve deepfake detection.

\subsection{KAZE Features}
The KAZE features are computed to capture the multi-scale and nonlinear structure of the keyframes. The KAZE algorithm involves detecting keypoints and computing descriptors. 
This process is involved by applying nonlinear diffusion filtering to the keyframe \(I \) to create a nonlinear scale space. The keypoints are detected with the KAZE detector using the formula: \(\{I(x_i,y_i)\}K_(i=1)\). 
For each keypoint \((x_i, y_i)\), compute a descriptor vector \(d_i\) that represents the local image patch around the keypoint. Concatenate the descriptor vectors into a single feature vector \(D\). If the total number of features exceeds a predefined length, the feature vector is truncated or padded as follows:
\[
D = [d_1, d_2, ..., d_k]_{(1:m)}
\] where \(m\) is the designed length of the KAZE feature vector. 
Finally, computed descriptors are used for each key point. KAZE descriptors are computed by sampling the responses of the nonlinear scale space at keypoint locations using orientation and scale information.
The extracted features are later used to classify the video as either fake or real. This classification is accomplished via ML classifiers with KAZE robustness in extracting image features to detect deepfakes precisely.

\subsection{Proposed Feature Fusion and Classification}
The process of combining LBPs and KAZE features for image classification involves extracting two distinct sets of features from a single image, merging these feature sets into a unified feature vector, and then using this combined vector to train a classifier. This procedure improves classification performance, particularly for detecting deepfake content by utilizing the included strengths of KAZE (keypoint detection and description) and LBP (texture analysis).

The proposed fusion model is a comprehensive method for image classification that integrates LBP or HOG and KAZE features, followed by the selected classifier. Initially, the algorithm extracts LBP or HOG features, which capture texture features with the distribution of binary patterns in the neighborhood of each pixel. Mathematical representations (Equations~\ref{eq:proposed_equation} and ~\ref{eq:proposed_equation_final}) by the LBP histogram are used to normalize and achieve uniform feature scaling. Concurrently, KAZE features are extracted by detecting key points and computing their descriptors, effectively capturing local invariant features. These descriptors are integrated into a single vector, which is then truncated or padded to maintain a consistent feature length. The LBP and KAZE feature vectors are combined to create a single feature representation for each image, utilizing the improved strengths of both feature extraction methods.

The process of merging the LBP and KAZE features is explained in these steps. First, we extract the LBP feature vector \(F_LBP\) from the histogram \(H_LBP\) by employing the following formula:
\begin{equation}
	\label{eq:proposed_equation}
	F_{\text{LBP}} =  H^{\text{LBP}}(0), H^{\text{LBP}}(1), \ldots, H^{\text{LBP}}(2^P - 1)
\end{equation}

Second, the KAZE feature vector \(F_{KAZE}\) is extracted from the concatenated descriptor vectors \(D\) by \(F_{KAZE} = D\). 

Finally, the \(F_{LBP}\) and \(F_{KAZE}\) feature vectors are concatenated as~\mbox{follows}:
\begin{equation}
	\label{eq:proposed_equation_final}
	\mathbf{F}_{\text{combined}} = [\mathbf{F}_{\text{LBP}}, \mathbf{F}_{\text{KAZE}}]
\end{equation}

The integration of LBP and KAZE features improves the algorithm's robustness and accuracy in deepfake classification tasks, especially when detecting false or real images. LBP features improve in detecting texture patterns, which are important in distinguishing between real and fake images because deepfakes often possess irregular or inconsistent textures. In contrast, KAZE features retrieve fine-grained keypoints, which is necessary for detecting minor modifications that may not greatly alter texture but have an impact on the structural integrity of the image. The proposed fusion model improves the classifier's capability to identify deepfakes by integrating these two feature sets to provide a deeper and more discriminative feature space. In experiments utilizing fake images, the improved detection capabilities and decreased false positives demonstrate how the integration of texture-based and keypoint-based features results in the improvement of classification accuracy. The proposed fusion model improves performance on deepfake detection tasks by utilizing the advantages of KAZE (robust keypoint descriptors) and LBP (sensitive to textures). Furthermore, Algorithm~\ref{al:fusion} is expressed to provide a more detailed explanation from the perspective of implementation.

\begin{algorithm}[H]
	\setstretch{1.15}
	\caption{Algorithm for merging LBP/HOG and KAZE features and classification}
	\label{al:fusion}
	\begin{algorithmic}[1]
		\Require Set of images $\{I_1, I_2, \ldots, I_N\}$, corresponding labels $\{y_1, y_2, \ldots, y_N\}$.
		\Ensure Classification accuracy.
		
		\Function{ExtractLBPFeatures}{$I$, $R=3$, $P=24$}
		\State Convert image $I$ to LBP and HOG using radius $R$ and $P$ points.
		\State \Return $\hat{H}_{LBP/HOG}$
		\EndFunction
		
		\Function{ExtractKAZEFeatures}{$I$, $M=64$}
		\State Detect keypoints in $I$ using KAZE.
		\State Compute descriptors for each keypoint and Concatenate descriptors into $\mathbf{D}$.
		\State \Return $\mathbf{D}$
		\EndFunction
		
		\Function{CombineFeatures}{$\mathbf{F}_{LBP/HOG}$, $\mathbf{F}_{KAZE}$}
		\State \Return $[\mathbf{F}_{LBP/HOG}, \mathbf{F}_{KAZE}]$
		\EndFunction
		
		\Function{PrepareDataset}{$\{P_1, P_2, \ldots, P_N\}$, $\{y_1, y_2, \ldots, y_N\}$}
		\For{each path $P_i$ in $\{P_1, P_2, \ldots, P_N\}$}
		\State Load image $I$ from path $P$ and convert it to grayscale.
		\State $\mathbf{F}_{LBP/HOG}^{(i)} \gets$ \Call{ExtractLBP/HOGFeatures}{$I_i$}
		\State $\mathbf{F}_{KAZE}^{(i)} \gets$ \Call{ExtractKAZEFeatures}{$I_i$}
		\State $\mathbf{F}_{combined}^{(i)} \gets$ \Call{CombineFeatures}{$\mathbf{F}_{LBP/HOG}^{(i)}$, $\mathbf{F}_{KAZE}^{(i)}$}
		\State Append $\mathbf{F}_{combined}^{(i)}$ to features list.
		\EndFor
		\State \Return feature matrix $\mathbf{X}$ and label vector $\mathbf{y}$
		\EndFunction
		
		\Function{TrainAndEvaluate}{$\mathbf{X}$, $\mathbf{y}$}
		\State Split \& Train classifier on the training set. Predict, Compute accuracy of predictions.
		\State \Return accuracy
		\EndFunction
		
		\State \textbf{Main Program}
		\State $\mathbf{X}, \mathbf{y} \gets$ \Call{PrepareDataset}{$\{P_1, P_2, \ldots, P_N\}$, $\{y_1, y_2, \ldots, y_N\}$}
		\State accuracy $\gets$ \Call{TrainAndEvaluate}{$\mathbf{X}, \mathbf{y}$}
		
	\end{algorithmic}
\end{algorithm}

\section{Implementation}\label{sec:exp-result}
\subsection{Experimental Design}
\label{sec:exp-setup}

This section summarizes the experimental details and results of deepfake datasets for FaceForensic++~\cite{roessler2019faceforensicspp} and Celeb-DF ~\cite{Li_2020_CVPR}. The robustness of the fusion (LBP, HOG, and KAZE) features under various classifiers is evaluated. We applied some preprocessing on the raw data prior to experimentation with the datasets. The complete video sequence is not taken into account. Instead, as described in Section \ref{sec:proposed-framework}, some keyframes are extracted. Both fake and real videos are included in the video dataset. After being extracted, the frames were placed in a folder with the appropriate name. The NVIDIA 3090 GPU was used for feature extraction and the development of ML algorithms.

\begin{figure}[H]
	\includegraphics[scale=0.99]{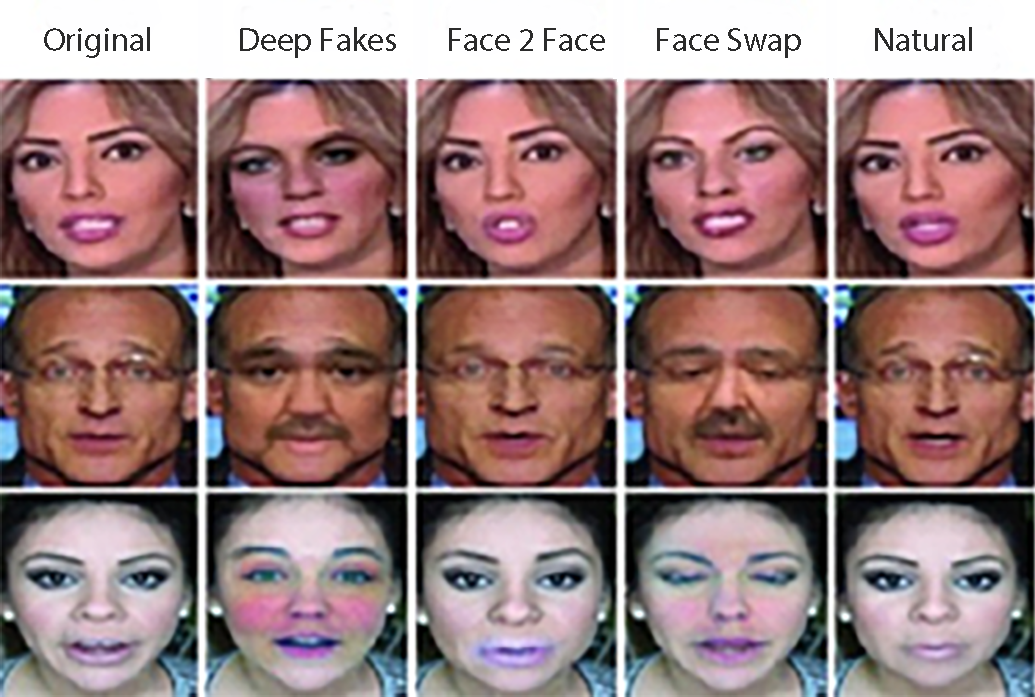}
	\caption{An example of fake faces from the FaceForensics++ dataset. The pristine image is in the first column, whereas the forged images produced by DeepFakes, Face2Face, FaceSwap, and NeuralTextures are in the second through fifth columns~\cite{roessler2019faceforensicspp}.} 
	\label{fig:dataset-sample}
\end{figure}

FaceForensics++~\cite{roessler2019faceforensicspp} is the popular publicly available forensic dataset that includes 1000 original video sequences that have been manipulated using four distinct face manipulation techniques: deepfake, Face2Face, FaceSwap, and NeuralTextures. The dataset consists of 977 YouTube videos with 48,431 face counts and a data size of 575 mb, all of which feature a clearly visible face, 
 allowing automated tampering methods to produce highly accurate forgeries (an example of the dataset is depicted in Figure~\ref{fig:dataset-sample}). We targeted deepfake videos and their original equivalents for our experiment. The dataset was created by extracting keyframes from several videos. Following preprocessing, there were 2946~fake images and 2930~real images in the training set. There were 198 real and 197 fake images in the validation.

The Celeb-DF~\cite{Li_2020_CVPR} dataset is divided into real and fake video/frames, where the real videos are 590~original YouTube videos with people of all ages and the fake videos are 5639 deepfakes. Keyframes from these videos were extracted to create the dataset, making sure that the images mostly featured the faces of different celebrities. The dataset consists of 100 genuine images, where 900 deepfake images make up the validation set and 1130 real images and 8022~deepfake images make up the training set. A balanced evaluation framework for deepfake detection models is provided by the test set, which consists of 340 deepfake images and 178~real~\mbox{images}.

There are two classes in the Celeb-DF dataset: real and fake. The percentage of fake classes is higher than that of real classes. The FaceForensics++ dataset has five classes, including 1000 videos in each class. The original video is the first class, and the other four classes are videos that have been altered/are fake. These five classes were reduced to binary classes in equal parts for the purposes of this study. To remove the dataset imbalance problem, a similar process~\cite{9563923} is adopted, where 800 videos were finalized from each dataset, 400 of which were chosen for each class. Following the average number of frames prepared for the experiment, the detailed dataset is displayed in Table \ref{tab:db}.

\begin{table}[H]
\caption{FaceForensics++ and Celeb-DF dataset compositions.} 
\label{tab:db}
\begin{tabularx}{\textwidth}{LCC}
\toprule
\textbf{Dataset}         & \textbf{Real Images} & \textbf{Fake Images} \\ \midrule
Celeb-DF        & 382  & 346  \\ 
FaceForensics++ & 496  & 458  \\ \bottomrule
\end{tabularx}
\end{table}

\subsection{Evaluation Criteria} 
Empirical benchmarking is a popular way to accurately analyze feature extraction and training times. This involves directly quantifying the time spent throughout experimental runs, resulting in exact and dependable data. This method is especially useful for machine learning tasks, where computational complexity varies depending on dataset size, hardware capabilities, and specific implementation choices. In this paper, we describe the methodological approach used to calculate feature extraction, training, and inference times for ML classifiers, providing a thorough assessment of computing efficiency.

The time required for feature extraction can be computed as follows:
for each feature extraction method, the extraction process for all the data points in the dataset is applied. The start and end times \(T_{feature} = T_{end} - T_{start}\) are recorded for each feature extraction run~\cite{hastie2009elements}. To mitigate variability owing to hardware or background processes, the feature extraction process is repeated multiple times, and the average time is computed as follows:
\begin{equation}
	\label{eq:4}
	Average Feature Extraction Time = \frac{\sum_{i=1}^{n} T_{\text{feature},i}}{n}
\end{equation}
where \(T{feature, i}\) denotes the feature extraction time for the \(i-th\) run and \(n\) is the number of runs. For reporting purposes, the time per data instance (e.g., per frame in video processing) can also be computed as follows:
\begin{equation}
	\label{eq:5}
	T_{feature, instance} = \frac{ T_{feature}}{N}
\end{equation}
where \(N\) is the total number of data instances. Training time refers to the duration required to train an ML model on a specified dataset. For each classifier (e.g., RF, SVM, and CNN), the training process is initiated and calculated by recording the start and end times as follows: \(T_{train} = T_{end} - T_{start}\)~\cite{fawzi2016robustness}. Similar to feature extraction, it is often beneficial to perform multiple runs and compute the average through the following equation to obtain a reliable~\mbox{estimate}:
\begin{equation}
	\label{eq:6}
	Average Training Time = \frac{\sum_{i=1}^{n} T_{\text{train},i}}{n}
\end{equation}

For larger datasets, the training time may also be approximated on the basis of model complexity. For example, the training time complexity of RF with \(N\) trees is generally \(O(N log N)\), whereas the support vector classifier may exhibit \(O(N_2)\) complexity. The inference time is the duration required to classify a new instance after training. The total inference time over a dataset \(T_{inference}\) can be approximated by \(T_{inference} = T_{inference, instance} x N\), where \(N\) is the total number of instances in the test dataset. Multiple test runs were carried out to establish temporal consistency among approaches, with start and end times carefully documented. Measuring the time necessary for each instance provides for more detailed comparisons and brings out performance differences more clearly. This established approach makes sure that all important time calculations for feature extraction, training, and inference are directly comparable, resulting in a rigorous and repeatable experimental framework.

\subsection{Results and Discussion}
The results are further provided and analyzed in depth. This study proposed a feature-level fusion of LBP, HOG, and KAZE features for classification via the FaceForeensics++ and Celeb-DF datasets. The evaluation of the results on the basis of the provided validation is presented below:
Table~\ref{tab:fusion-LBP-KAZE-features} presents the classification accuracies obtained using various classifiers with various feature sets, including LBP alone, KAZE alone, and the fusion of LBP, HOG, and KAZE features. The experiment was conducted on the FaceForensics++ and Celeb-DF datasets to evaluate the effectiveness of these features in distinguishing between genuine and manipulated visual content/deepfake.

\begin{table}[H]
	\caption{Classification accuracy of fusion of LBP, HOG, and KAZE features with FaceForensic++.}
	\label{tab:fusion-LBP-KAZE-features}
	\begin{tabularx}{\textwidth}{LLC}
		\toprule
		\multicolumn{2}{c}{\textbf{Fusion of Features with Classifiers}}  & \textbf{Accuracy} \\ \midrule
		{\multirow{2.5}{*}{LBP Features}} & Extra Trees Classifier & 71.22\%   \\ \cmidrule{2-3}
		                              & RF Classifier & 70.76\%   \\ \midrule
		                              		
		{\multirow{5}{*}{KAZE Feature}} & Extra Trees Classifier & 85\%   \\ \cmidrule{2-3}
		                              & Support Vector Classifier & 86.12\%   \\ \cmidrule{2-3}
		                              & RF Classifier & 75.70\%   \\ \cmidrule{2-3}
		                              & XGB Classifier & 65.29\%   \\ \midrule

		{\multirow{5}{*}{HOG + KAZE Feature}}& Extra Trees Classifier & 91\%   \\ \cmidrule{2-3}
		                              & Support Vector Classifier & 92.12\%   \\ \cmidrule{2-3}
		                              & RF Classifier & 85.70\%   \\ \cmidrule{2-3}
		                              & XGB Classifier & 83.19\%   \\ \midrule
        
		{\multirow{5}{*}{LBP + KAZE Features}} & Extra Trees Classifier & 82.61\%   \\ \cmidrule{2-3}
		                              & Support Vector Classifier & 86.22\%   \\ \cmidrule{2-3}
		                              & RF Classifier & 85.54\%   \\ \cmidrule{2-3}
		                              & XGB  Classifier & 88.56\%   \\ \bottomrule
		
	\end{tabularx}
\end{table}

Analyzing the results shows that both LBP and KAZE features individually perform well across different classifiers when tested on the FaceForensics++ dataset. LBP features demonstrated their effectiveness in texture-based analysis, achieving an accuracy of 71.22\% with Extra Trees and 70.76\% with Random Forest classifiers. Similarly, KAZE features, which focus on detecting structural alterations and keypoint variations in images, produced accuracies ranging from 75.70\% with Random Forest to 86.12\% with a Support Vector~\mbox{Classifier}.

HOG features also showed strong performance, with accuracies between 85.76\% and 92.12\% using a Support Vector Classifier. This result is close to the benchmark accuracy of 94.44\% for the FaceForensics++ dataset, which was achieved by EfficientNet~\cite{9412711}. These findings highlight the potential of integrating feature extraction techniques to improve deepfake detection performance.

\begin{figure}[H]
	\includegraphics[scale=0.8]{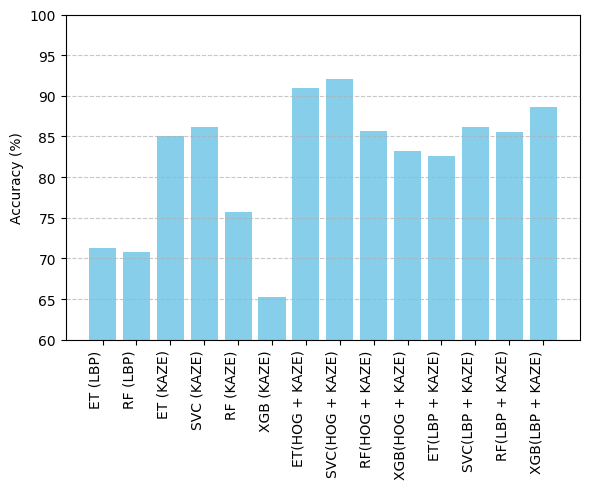}
	\caption{Outcome of the proposed and ML algorithms in terms of accuracy (i.e., RF, extra trees, and~SVC).}
	\label{fig:result-histogram}
\end{figure}

However, the most notable results were achieved by the fusion of HOG and KAZE features, demonstrating a clear advantage over individual feature sets. This fused approach showed superior performance across all classifiers, with accuracy rates of 91.12\% using Extra Trees, 92.12\% with the Support Vector Classifier, and an impressive 94.44\% when tested with the state-of-the-art EfficientNet. These results indicate that integrating texture-based (i.e., HOG) and keypoint-based (i.e., KAZE) features significantly improved the model's ability to detect deepfake content. These features allow for more effective detection of deepfakes compared to using either method alone (see Figure~\ref{fig:result-histogram} for a detailed visualization).

Future research could focus on strengthening the fusion technique of HOG and KAZE features to improve feature selection and reduce dimensionality. Another promising direction is the exploration of DL architectures, such as CNNs, to gain deeper insights into hierarchical feature representations and further boost classification accuracy. Additionally, evaluating the proposed fusion model on larger and more diverse datasets other than FaceForensics++ would help assess its robustness in real-world scenarios. These advancements would play an important role in strengthening deepfake detection, particularly in addressing the growing challenges in digital manipulation and deepfake technologies.

The results presented in Table \ref{tab:fusion-LBP-KAZE-features-celeb-df} highlight the classification accuracy of different feature extraction techniques when applied to deepfake detection using the Celeb-DF dataset. The findings demonstrate that combining multiple feature descriptors improves detection performance compared to using individual features alone. Among the single-feature approaches, LBP features with a Support Vector Classifier (SVC) achieved the highest accuracy (72\%), which yielded a lower accuracy of 68\%. This suggests that LBP is more effective in capturing local texture variations relevant to distinguishing real from fake images. When feature fusion was applied, the combination of HOG and KAZE features achieved the highest accuracy (78\%), showing a significant improvement over individual features. This indicates that integrating gradient-based descriptors (HOG) with keypoint-based features (KAZE) provides a more comprehensive representation of image characteristics, improving classification reliability. Similarly, LBP combined with KAZE achieved an accuracy of 75\%, further confirming that KAZE features contribute positively to deepfake detection by enhancing feature diversity.

\begin{table}[H]
	\caption{Classification accuracy of fusion of LBP, HOG, and KAZE features with Celeb-DF.}
	\label{tab:fusion-LBP-KAZE-features-celeb-df}
	\begin{tabularx}{\textwidth}{LLC}
		\toprule
		\multicolumn{2}{c}{\textbf{Fusion of Features with Classifier}}  & \textbf{Accuracy} \\ \midrule
		LBP Features & Support Vector Classifier & 72\%   
		\\ \midrule
		HOG Features & Support Vector Classifier & 68\%   
		\\ \midrule
		HOG + KAZE Features & Support Vector Classifier & 78\%   
		\\ \midrule
		LBP + KAZE Features & Support Vector Classifier & 75\%   
		\\ \bottomrule		
	\end{tabularx}
\end{table}

Overall, these results demonstrate that fusion strategies, particularly HOG + KAZE, are more effective than single-feature approaches in deepfake detection on the Celeb-DF dataset. The findings suggest that feature extraction techniques can improve the robustness of detection models, making them more resilient to sophisticated deepfake manipulations. Future research might investigate refining feature selection and classifier tweaking to improve overall performance.


Table \ref{tab:time-comparison} presents a comparison of execution times for feature extraction and classification, measured in milliseconds on both GPU and CPU. This comparison evaluates the inference time and classification accuracy of different methods, as outlined in the evaluation criteria (Section \ref{sec:exp-setup}), using the FaceForensics++ dataset.

\begin{table}[H]
\tablesize{\small}
	\caption{Execution time comparison for feature extraction and classification.}
	\label{tab:time-comparison}
	\begin{tabularx}{\textwidth}{lcccC}
		\toprule
		\textbf{Methods} & \textbf{Feature Extraction} & \textbf{Training} & \textbf{Inference GPU} & \textbf{CPU}\\ \midrule
		Random forest                & 0.5 s   & 30 m  & 15 ms & 92 ms  \\
		Extra Trees Classifier       & 0.5 s   & 25 m  & 15 ms & 95 ms \\
		Support Vector Classifier    & 0.5 s   & 60 m  & 13 ms & 63 ms\\
		XGB Classifier               & 0.5 s   & 45 m  & 12 ms & 75 ms\\
		Support Vector Machine       & 0.5 s   & 120 m & 25 ms & 85 ms\\
        XceptionNet                  & 0.5 s   & 210 m & 20 ms & 2  s\\
		Convolutional Neural Network & 0.5 s   & 180 m & 10 ms & 1.5 s\\
		HOG + KAZE (Proposed)        & 1.0 s   & 45 m  & 16 ms & 67 ms\\
        LBP + KAZE (Proposed)        & 0.5 s   & 30 m  & 09 ms & 56 ms\\		
		\bottomrule
	\end{tabularx}
\end{table}

Feature extraction time reflects the duration required to extract LBP, HOG, and KAZE features, in contrast with the extraction times of traditional machine learning models. Training time refers to the time duration needed to train the classifier after feature extraction, while inference time measures the time taken to classify a single instance once the model has been trained. This analysis provides insights into the computational efficiency of various approaches, helping to assess their suitability for real-time deepfake detection.

These tables provide a clear comparison of the efficiency and effectiveness of the proposed fusion model using KAZE and HOG features against traditional ML approaches. Table~\ref{tab:time-comparison} demonstrates that while feature extraction with KAZE and HOG may take slightly longer (per frame) than traditional ML models on GPU, it is significantly faster than training DL models. Table~\ref{tab:fusion-LBP-KAZE-features} highlights that the proposed fusion model achieves competitive accuracy while maintaining comparative training and inference time, making it suitable for real-time applications where speed is important (see Table \ref{tab:time-comparison} for GPU and CPU inference). These comparisons underscore the practical advantages of the proposed fusion model in terms of computational efficiency without compromising classification performance. The proposed fusion-based methods perform better in terms of training time, despite CNN's marginally better performance and accuracy, but are lacking in inference time, especially when the CPU is used. Although CNN's inference is somewhat superior when the GPU is used compared to the proposed fusion model, the methods are reliable and perfect for resource-constrained situations, as they performed better when the CPU is used. 
 The fusion of HOG and KAZE provides a good balance between computational efficiency and classification performance.

\subsection{Future Work and Implications of Visual Information Security}
The fusion of LBP, HOG, and KAZE features has proven effective in detecting deepfake content. HOG captures texture patterns, while KAZE detects structural distortions introduced by deepfake generation techniques. Future research could refine classifiers to improve the model’s ability to distinguish between real and fake content. Exploring advanced hybrid techniques like ORB, DSIFT, and Wavelet Transform Features could further improve detection accuracy and computational efficiency. Additionally, dimensionality reduction methods such as PCA and t-SNE can optimize feature selection, while DL approaches like hybrid CNN architectures or GANs could bolster the robustness of deepfake~\mbox{detection}.

This approach has strong potential for forensic and legal applications, providing a reliable means to verify the authenticity of digital media in critical legal proceedings. It could also contribute to real-time authentication systems for digital media platforms, potentially integrating blockchain or watermarking techniques for added security. The research highlights the importance of robust feature extraction methods in deepfakes, and future efforts should focus on refining these techniques and adapting to new manipulation strategies to ensure continued efficacy in securing digital media.

\section{Conclusions}
\label{sec:conclusion}
In conclusion, this study introduced an effective approach for detecting deepfake images utilizing texture-based features through the fusion of HOG/LBP and KAZE within an ML framework. The computational load is significantly reduced compared to traditional DL models, making this method ideal for real-time applications with limited processing resources.  The experiments using classifiers such as RF, XGBoost, extra trees, and support vector classifiers demonstrated the distinct advantages of each method in evaluating feature importance across HOG feature bands. The feature-level fusion technique further improved performance on both the FaceForensics++ and Celeb-DF datasets, achieving an accuracy of 92.12\% and 78\%, respectively. This approach not only improves accuracy and efficiency in detecting deepfake content but also provides a scalable solution against the potential abuse of technology and its consequences on politics, finance, and personal privacy. Beyond deepfake detection, the method holds the potential for authenticating various forms of digital content, emphasizing its broad applicability in fields that require reliable visual data~\mbox{verification}.

\vspace{6pt}

\funding{This study was financially supported by the Seoul National University of Science and Technology, Seoul, South Korea.}

\conflictsofinterest{The authors declare no conflicts of interest.}

\begin{adjustwidth}{-\extralength}{0cm}

\reftitle{References}

\PublishersNote{}
\end{adjustwidth}
\end{document}